\pdfoutput=1

\documentclass[11pt]{article}

\usepackage[]{acl}

\usepackage{times}
\usepackage{latexsym}
\usepackage{amssymb}
\usepackage{bbding}
\usepackage{float}
\usepackage[T1]{fontenc}

\usepackage[utf8]{inputenc}

\usepackage{microtype}
\usepackage{graphicx}
\usepackage{soul}

\title{McQueen: a Benchmark for Multimodal Conversational Query Rewrite }

\author{Yifei Yuan$\mathbf{^1}$\thanks{\enspace Work done when Yifei Yuan was an intern at Alibaba. This work was supported by Alibaba Group through Alibaba Research Intern Program, and a grant from the Research Grant Council of the Hong Kong Special Administrative Region, China (Project Codes: 14200620).}, Chen Shi$\mathbf{^2}$, Runze Wang$\mathbf{^2}$,\\
\textbf{Liyi Chen$\mathbf{^3}$, Feijun Jiang$\mathbf{^2}$, Yuan You$\mathbf{^2}$, and Wai Lam$\mathbf{^1}$} \\
        $^1$The Chinese University of Hong Kong \\ $^2$Alibaba Group \\
        $^3$Nankai University \\
        \texttt{\{yfyuan,wlam\}@se.cuhk.edu.hk} \\
        \texttt{\{deling.sc,yunze.wrz,feijun.jiangfj,youyuan.yy\}@alibaba-inc.com} \\
        \texttt{liyichen@mail.nankai.edu.cn}}

\begin{document}
\maketitle

\begin{abstract}
The task of query rewrite aims to convert an in-context query to its fully-specified version where ellipsis and coreference are completed and referred-back according to the history context. Although much progress has been made, less efforts have been paid to real scenario conversations that  involve drawing information from 
more than one modalities. In this paper, we propose the task of multimodal conversational query rewrite (McQR), which  performs query rewrite under the multimodal visual conversation setting. We collect a large-scale dataset named McQueen based on manual annotation, which contains 15k visual conversations and over 80k queries where each one is associated with a fully-specified rewrite version. In addition, for entities appearing in the rewrite, we provide the corresponding image box annotation. We then use the McQueen dataset to benchmark a state-of-the-art method for effectively tackling the McQR task, which is based on a multimodal pre-trained model with pointer generator. Extensive experiments are performed to demonstrate the effectiveness of our model on this task\footnote{The dataset and code of this paper are both available in \url{https://github.com/yfyuan01/MQR}}.

\end{abstract}

\section{Introduction}
Recent years have witnessed an increasing attention in conversational-related tasks, such as conversational question answering~\cite{Choi2018QuACQA,Reddy2019CoQAAC}, visual conversation  modeling~\cite{Das2017VisualD}, etc. One main challenge in multi-turn conversation modeling is that information from context history is easy to be abbreviated or omitted in the follow-up queries, causing the so-called coreference and ellipsis. To address this concern, the task of query rewrite~\cite{Elgohary2019CanYU,Pan2019ImprovingOD,Su2019ImprovingMD} aims to reconstruct the original query to a fully specified form based on its history context. The rewrite eliminates the coreference and ellipsis in the original query without changing its semantic information, thus helping turn the more challenging multi-turn conversation modeling problem to a single-turn version. 

\begin{figure}
    \centering
    \includegraphics[width=\linewidth]{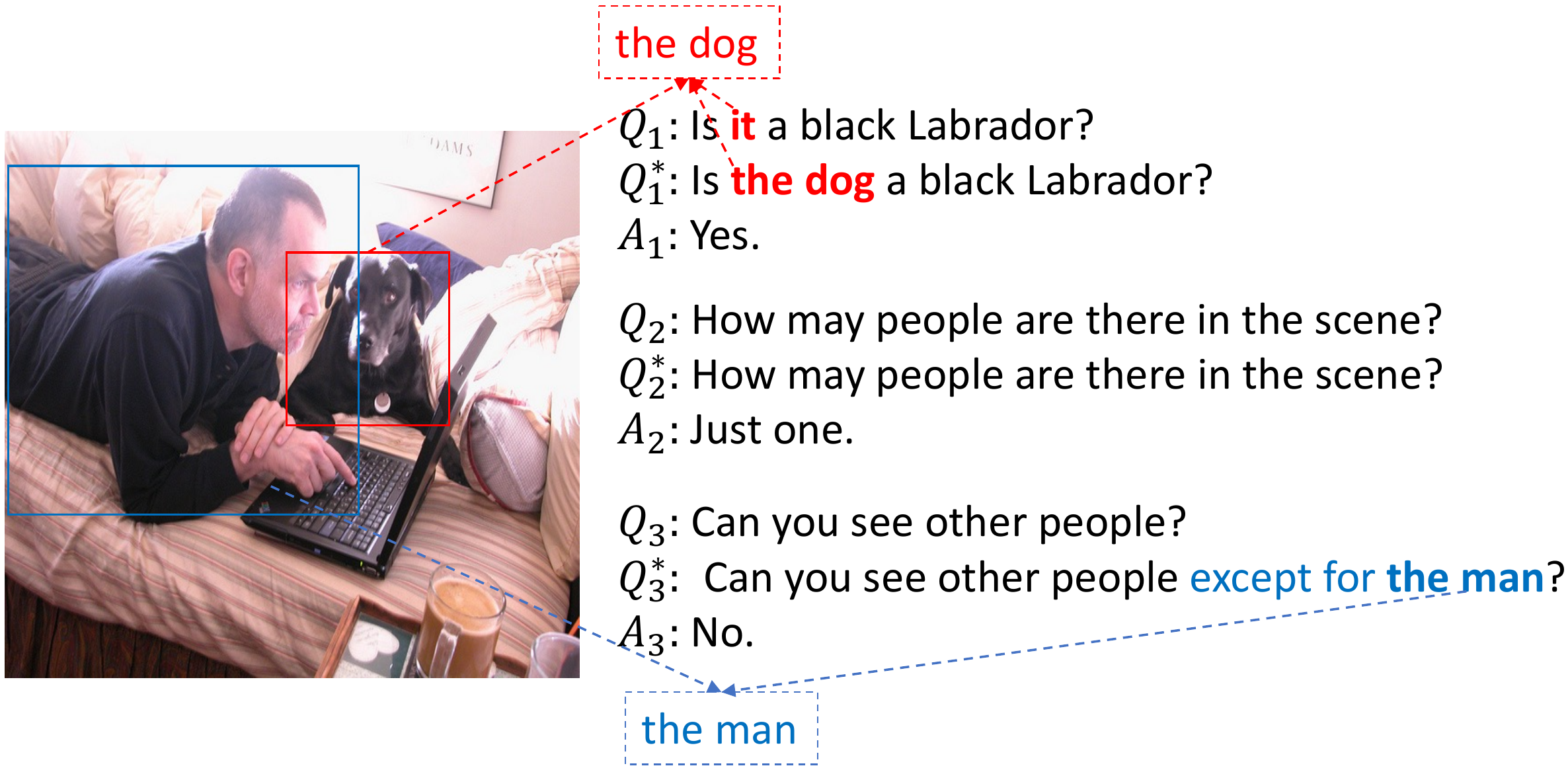}
    \caption{An example of the mulitmodal query rewrite task, where  $\mathcal{Q}$, $\mathcal{Q^*}$, and $\mathcal{A}$ denote the queries, their corresponding rewrites and the answers. Red color denotes the coreference rewrite part and blue denotes the ellipsis rewrite part.  Image boxes are utilized  for representing the area of the entities in the rewrite. }
    \label{graphic1}
\end{figure}
Following this line, several attempts have been made in the query rewrite task which achieve decent performance on the natural language level. Nevertheless, conversations in real scenario tend to involve knowledge from more than one modalities, such as vision, text, speech, etc. Information from different modalities is not handled in isolation, but often integrated together to improve the quality of perception and understanding. For example, as shown in Figure \ref{graphic1}, in the first turn of the visual conversation, with the lack of context, the user directly uses the pronoun ``it'' to refer to the dog in the image. In the third turn, for ellipsis that does not appear in the context history, in order to perform ellipsis completion, one also needs to find clues from the corresponding image. Rewriting the query under the circumstance where grounding outside the text information is needed poses more challenges to traditional query rewrite models that based only on textual features.

In this paper, we propose the task of multimodal conversational query rewrite (McQR), which aims to perform query rewrite under the multimodal visual conversation setting. To achieve this goal, we collect a large-scale dataset called McQueen. Specifically, for each visual conversation consisting of an image and the corresponding history question-answer context, we provide manual rewrite for the query, where the coreference resolution and ellipsis completion are performed respectively. Furthermore, in order to assist downstream tasks such as coreference entity detection, for all the entities appearing in the rewrite, we annotate the image boxes for representing their corresponding image area.


We then use the McQueen dataset to benchmark a state-of-the-art method for effectively tackling the McQR task. Inspired by the big success of pre-trained models such as BERT~\cite{Devlin2019BERTPO}, our model is based on a multimodal pre-trained model where interactions between different modalities can be better captured. Furthermore, we enhance the model with a pointer generator specially designed for the multimodal  Transformer blocks~\cite{Vaswani2017AttentionIA},  so that the rewritten query is either generated from scratch or copied from certain contextual  parts with high attention weights. Extensive experiments are conducted to compare our method with several state-of-the-art methods. Our model outperforms all the methods in both the McQR task and two subtasks. We further perform analysis to investigate the role of different modalities in this task and demonstrate that the introduction of image information provides extra guidance for the concerned query rewrite task.

In summary, the contribution of our paper lies in three folds:
\begin{itemize}
    \item We formally define the task of multimodal conversational query rewrite (McQR), which aims to generate a fully-specified rewrite query based on both the context history and the visual image.
    \item We propose a large-scale dataset McQueen, containing 15k visual conversations and over 80k rewrites. For the entities appearing in the rewrites, we also annotate the image boxes for representing their corresponding image area. 
    \item  We benchmark a multimodal Transformer-based model with pointer mechanism for effectively tackling the McQR task. Extensive analysis shows the role of different modalities in our model.
\end{itemize}

\section{Related Work}
\subsection{Query Rewrite}
The task of query rewrite provides reconstructed queries based on abbreviated in-context queries without changing their semantic meaning. First introduced by ~\cite{Elgohary2019CanYU,Su2019ImprovingMD,Pan2019ImprovingOD}, most works formulate it as a standard generation task, which can be solved via a Sequence-to-Sequence model~\cite{Quan2019GECORAE,Vakulenko2021QuestionRF,Anantha2021OpenDomainQA}. Some attempts have been made to introduce a multi-task learning setup in order to enhance the training process~\cite{Rastogi2019ScalingMD,Song2020ATC,Zhang2020FillingCE}. Some works seek to focus on query rewrite under the  low-resource scenario ~\cite{Yu2020FewShotGC,Voskarides2020QueryRF,Yu2021FewShotCD}. For modeling the linguistic knowledge in conversational context more effectively, prior knowledge is leveraged such as using semantic role labeling to provide extra guidance ~\cite{Xu2020SemanticRL}, and reducing the generation search space via sequence tagging ~\cite{Hao2021RASTDD}. 
Although these works have achieved great performance on their corresponding task, query rewrite under the multimodal setting has not been explored.

\subsection{Visual Coreference Resolution}
Visual dialog entails answering a set of questions grounded by an image~\cite{Das2017VisualD}. Based on that, visual coreference resolution involves linking the words in the text (usually nouns and pronouns) to a certain area in the image~\cite{Kong2014WhatAY,Kottur2018VisualCR}. Following this line, ~\citet{Li2021ModelingCR} restrict coreference resolution to pronouns and resolve coreferences in visual dialog in an unsupervised way. ~\citet{Yu2019WhatYS} define the task of visual pronoun coreference resolution where a dataset called VisPro and a model called VisCoref are benchmarked accordingly. Based on that,  ~\citet{Yu2022VDPCRIV} resolve pronoun coreference and propose a novel framework to improve visual dialog understanding. This task can be seen as a subtask of the McQR task where coreference resolution and ellipsis completion are both taken into account.

\section{The McQueen Dataset}

\subsection{Dataset Overview}
Our dataset is based on a visual dialog dataset called VisDial~\cite{Das2017VisualD}. The original VisDial dataset consists of over 133k dialogs, each associated with an image and 10 rounds of question-answer pairs. All  question-answer pairs are conducted in a conversational format and revolve around the content of the picture.

Our dataset randomly selects 15k conversations from the VisDial dataset with the total of over 80k rewrite utterances. For each query in a visual conversation, we conduct manual annotation to resolve the information omission. The query is reconstructed based on the image as well as the history context so that the coreference and ellipsis are referred-back or completed. For negative queries that do not contain any information omission, the rewrite stays the same as the original query. In addition, for all the entities appearing in the coreference and ellipsis, we annotate the   image boxes for representing their corresponding image areas.

\subsection{Dataset Construction}
\subsubsection{Text Rewrite Annotation}
For manual annotation, we hire 16 annotators in total. Before the annotation starts, we provide 100 examples for all the annotators to refer to. We also provide a guideline and some tutorials by listing some typical coreference and ellipsis cases so that the bias and language style shift between individuals are minimized as possible. After that, the annotators start working on a small portion of data where query rewrite is performed. After all the results are returned and the data quality is checked, the main annotation phase begins and the rest of data is labeled. On average, each annotator is in charge with the rewrite of 5059 queries. The rewrite annotation interface can be seen in Appendix \ref{interface}.
\subsubsection{Image Box Annotation}
Besides the rewrite annotation, we also provide image annotation to assist downstream or related tasks (e.g. coreference entity detection). The image box annotation begins right after the rewrite annotation. The overall procedure also follows the (1) tutorial (2) trial phase (3) main phase pipeline. Specifically, the annotators have to extract the entities in the ellipsis and coreference part and draw the bounding boxes of them in the image. Each annotator is in charge of the image annotation of the rewrites written by him/herself. The image annotation interface can be seen in Appendix  \ref{iba}.
\subsubsection{Quality Control}
After the all the annotation is finished, we re-group and shuffle the annotators to perform cross quality inspection. Each group is asked to check the annotation results of other groups. In addition, two new annotators who do not take part in the annotation phase are recruited to check the quality of all the annotation results. The annotators have to answer three questions for each query : (1) \textit{Is the rewrite result  correct or not?} (2) \textit{Are all the coreference and ellipsis resolved in the rewrite?} (3) \textit{Are the entities in the coreference and ellipsis correctly annotated in the image?} All the conversation rewrites must get the all ``yes'' result from all the annotators before official acceptance, otherwise they are collected to be revised and re-checked (the questionnaire interface is shown in Appendix \ref{qc}). The whole check-revise process lasts for three iterations. Considering chance agreement, we measured the Inter-Annotator Agreement (IAA) in terms of Cohen’s $\kappa$~\cite{Cohen1960ACO}. The final $\kappa$ score is 0.82, reaching the ``almost perfect'' level\footnote{According to ~\cite{Landis1977TheMO}, if $\kappa>=0.81$}. Besides, after each quality check iteration, we randomly sample 100 conversations from the dataset and manually evaluate the utterance-level precision and recall rate, where precision denotes the rate of the retrieved rewrites being correct, while the recall rate records the portion that the coreference and ellipsis being handled. The precision and recall rate in the 1st/2nd/3rd iteration are (89.0\%, 87.1\%)/(95.5\%, 94.2\%)/(98.3\%, 98.2\%), respectively.
\subsubsection{Annotation Cost and Duration}
The overall phase including the annotation and quality check spanned for 10 weeks (from March to May 2022), where the annotation guidance lasts for 2 weeks, data annotation lasts for 5 weeks, quality check lasts for 3 weeks. All the annotators are English native speakers recruited from a professional data management company Appen\footnote{https://appen.com/}. The annotation costs \$5942 US dollars in total, with \$0.31 per utterance rewrite, \$0.03 per image box annotation. 
\subsection{Dataset Statistics}
\begin{table*}[]
    \centering
    \small
    \begin{tabular}{cccccccc}
    \hline
         & Task & Conv. Size & Utterance Size.& Modality & Coref. & Ellip. & Image Box\\
    \hline
         CANARD~\shortcite{Elgohary2019CanYU}& Query Rewrite & 5644 & 40527 &t &\Checkmark & \Checkmark & \XSolidBrush\\
         REWRITE~\shortcite{Su2019ImprovingMD}&Query Rewrite& 2000 & 40000 &t &\Checkmark & \Checkmark & \XSolidBrush\\
         RESTORE~\shortcite{Pan2019ImprovingOD}&Query Rewrite&  203965 & 203965 &t&\Checkmark&\Checkmark& \XSolidBrush\\
         VisDial~\shortcite{Das2017VisualD}& Visual Dialog& 133351 & 1333510 &t\&v & \XSolidBrush & \XSolidBrush& \XSolidBrush\\
         VisPro~\shortcite{Yu2019WhatYS}&VCR& 5000 & 29722 &t\&v&\Checkmark&\Checkmark&\XSolidBrush\\
         McQueen&McQR & 15000 & 80944 & t\&v &\Checkmark&\Checkmark&\Checkmark\\
    \hline
    \end{tabular}
    \caption{Comparison with existing datasets, where VCR denotes visual pronoun coreference resolution.}
    \label{tab:comparison}
\end{table*}
\begin{table}[t]
    \centering
    \small
    \begin{tabular}{c|c}
    \hline
        Characteristic  & \% Ratio \\
    \hline
        Queries w/ Coreference & 57.2 \\
        Queries w/ Ellipsis & 30.4 \\
        Queries w/o both & 13.5 \\
    \hline
    \end{tabular}
    \caption{Proportion of queries with coreference and ellipsis in our dataset.}
    \label{tab:proportion}
\end{table}
\begin{table}[]
    \centering
    \small
    \begin{tabular}{p{5.6cm}p{1.2cm}}
    \hline
    Avg. Num. of Turn per Conv.: & 5.40 \\
    Avg. Num. of Entity per Conv.: & 3.04 \\
    Avg. Num. of Image Box per Conv.: & 2.02 \\
    Avg. Length of Rewrites: & 6.65\\
    Avg. Length of Contexts: & 42.13\\
    
    \hline
    \end{tabular}
    \caption{Statistics of the dataset. The length is calculated by the number of English words.}
    \label{tab:dataset}
\end{table}
\begin{table}[]
    \centering
    \small
    \begin{tabular}{cccccc}
    \hline
         0-2 & 3-4 &5-6 &7-8 & 9+ &o/a\\
         15143 & 17064 &16431 & 16126 & 16180& 80944\\
    \hline
    \end{tabular}
    \caption{Number of rewrites concerning different history turn lengths in McQueen.}
    \label{tbl:turns}
\end{table}


According to Table \ref{tab:proportion}, 86.5\% of our dataset covers positive rewrite cases that coreference or ellipsis occurs in the query. Table \ref{tab:dataset} lists the  statistics of our dataset, where each visual  conversation contains 5.40 Q-A turns with 2.02 image boxes on average. Table \ref{tbl:turns} lists the number of rewrites in our dataset under different history context lengths, where most of the data contains context from 3-4 turns, and over 16k data contains context over 9 turns. 

Futhermore, we compare our dataset with existing datasets from related works. According to Table \ref{tab:comparison}, our dataset is: (1) \emph{more complete} - we provide manual annotation on both image and text levels, where the rewrite query and entity boxes are both provided; (2) \emph{more diverse} - since our dataset is designed not only for all the coreference cases, but also performs ellipsis completion in the rewrite; (3) \emph{much larger} - compared with existing VCR dataset, the dataset has a larger scale.
\begin{figure*}[t]
    \centering
    \includegraphics[width=\linewidth,height=55mm]{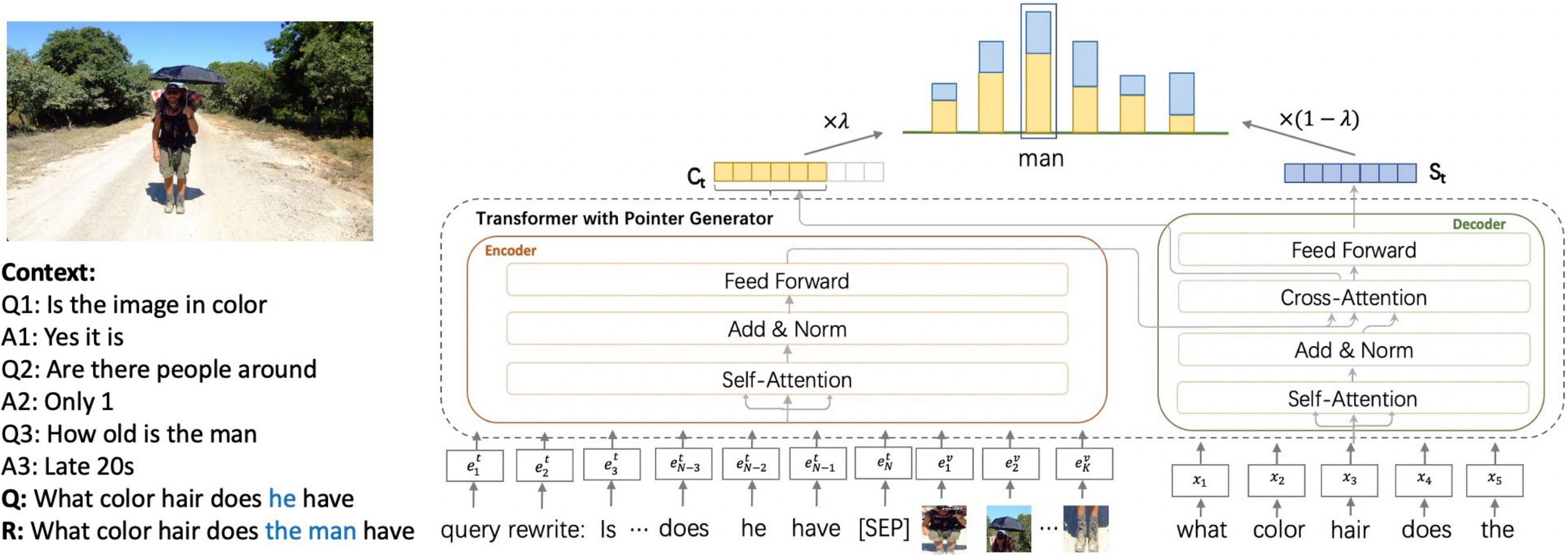}
    \caption{The overall architecture of our model.}
    \label{fig:model}
\end{figure*}
\section{McQR Task}
\subsection{Task Definition}
We define the task of multimodal conversational  query rewrite (McQR). Concerning a multimodal visual conversation with an image $I$ and the corresponding text conversation $D$, at time step $m$, given the conversation history $H$ with $m-1$ turns, the $i$-th turn consists of a query $q_i$ and an answer $a_i$. The goal of the task is to generate a rewrite query $r_m$ for the current query $q_m$ based on both image $I$ and textual history $H$, which can be represented as
\begin{equation}
    r_m = Rewriter(H,I,q_m)
\end{equation}

\subsection{Methods}
\label{model}
\subsubsection{Model Overview}
Figure \ref{fig:model} depicts the overall structure of our proposed model. Our model is based on a multimodal pre-trained model VL-T5~\cite{cho2021vlt5}, where a Transformer-based encoder and decoder are jointly trained to perform generation. Besides, we add a  pointer generator to the multimodal Transformer decoder so that the rewrite is either generated from scratch or copied from history context.
\subsubsection{Image and Text Embeddings}
\label{Encoding}
\textbf{Text Embedding}
The text input of our model includes two parts: the task prefix and the history context. The task prefix is a short prompt which aims to differentiate the concerned McQR task from other tasks in the pre-training phase (e.g. visual grounding). Specifically, we use ``query rewrite:'' as the task prefix. Following it, the history context contains all the utterances including all the queries and answers in previous turns. The input text is then tokenized and embedded before passed into the encoder. Following the setting in T5~\cite{Raffel2020ExploringTL}, we also add a relative position bias to each self-attention layer. As a result, the text input is  represented as  $e^t$.

\noindent\textbf{Image Embedding}
To extract image features, we first detect several object regions from the image, denoted as Region of Interest (ROI). Following previous works~\cite{cho2021vlt5,Lu2019ViLBERTPT}, we also use Faster R-CNN~\cite{Ren2015FasterRT} pre-trained on the Visual Genome dataset~\cite{Krishna2016VisualGC} to extract ROI features in our task. For each image $I$, we extract $n=36$ ROIs from it. The final visual vector can be represented as $e^v$.

\subsubsection{Encoder-Decoder Structure}
\textbf{Encoder.} We use a multimodal Transformer encoder-decoder structure to encode the image text features and generate the target rewrite. The multimodal encoder is a stack of $L$ Transformers which take the image and text representations as input
\begin{equation}
h_0 = [e^t_{1},..,e^t_{m},e^v_{1},...,e^v_{n}]
\end{equation}
\begin{equation}
h_i = FNN(MultiHead(h_{i-1},h_{i-1},h_{i-1}))
\end{equation}
\textbf{Decoder.} Similarly, the decoder is also a stack of $L$ Transformers. Given the decoder input $x_t$, the decoding step consists three phases, where the first sub-layer is a self-attention layer which can be represented as 
\begin{equation}
    d_i = MultiHead(d_{i-1},d_{i-1},d_{i-1})
\end{equation}
where $d_0=x_t$. After that, the second sub-layer calculates the cross attention between the encoder outputs and the decoder self-attention result
\begin{equation}
    s_i = MultiHead(d_i,h_L,h_L)
\end{equation}
The  third  sub-layer  is  a  position-wise  fully  connected feed-forward neural network, followed by a softmax layer to output the final probability
\begin{equation}
    P_{vocab}(y_t) =\sum_{i:w_i\in V}\beta_{t,i}= Softmax(FNN(s_i))
\end{equation}
where $\beta$ is the softmax score over the whole vocabulary $V$.
\begin{table*}[]
    \centering
    \small
    \begin{tabular}{p{2.8cm}cccccc}
    \hline
       &BLEU-2 &  BLEU-4 & ROUGE-2 & ROUGE-L & METEOR & EM \\
       Original  &57.27 & 48.44 & 54.86 &73.45 & 38.28 & - | - \\
       AllenNLP Coref & 59.72 & 50.21 & 56.09 & 76.07 & 39.87 & 8.97 | 96.14\\
    \hline
       L-Gen & 77.14 & 66.31 & 74.44 & 84.68 & 46.09 & 41.84 | 57.77\\
       T(E)-Gen & 78.03 & 65.40 & 75.53 & 86.80 & 47.32 & 41.37 | 56.58\\
       T(L)-Gen & 79.30& 67.16 &  76.28 & 88.14 &49.23 & 42.06 | 58.25\\
       L-Ptr & 77.46 & 69.46 & 76.07 & 86.65 & 48.63 & 47.87 | 66.72\\
       T(E)-Ptr & 79.52 & 68.26 & 77.27 & 88.49 & 50.10 &49.49 | 68.94\\
       T(L)-Ptr & 80.12 & 68.41 &78.40 & 89.54 & 50.56 & 48.15 | 69.53\\
    \hline
    VLBart & 90.01 & 84.28  & 88.67  &  93.80 & 58.87 & 64.28 | 86.89 \\
    VLT5 & 90.37 & 84.87 & 89.23 & 94.10 & 59.45 & 65.62 | 89.95 \\
    VLBart-Ptr & 90.16 & 84.52 & 88.87 & 93.89 & 59.26 & 64.87 | 87.22 \\
    VLT5-Ptr &\textbf{90.47} & \textbf{84.94} & \textbf{89.32} & \textbf{94.45} & \textbf{59.46}& \textbf{65.67} | \textbf{90.22} \\
    \hline
    \end{tabular}
    \caption{The main experiment on our dataset, where T(E) denotes Transformer with Early Fusion, T(L) denotes Transformer with Late Fusion. For EM rate, we report both EM for positive and negative rewrite samples.}
    \label{tab:mainexp}
\end{table*}

\noindent\textbf{Multimodal Transformer with Pointer Generator.} Additionally, following the motivation that most part of the rewrite sentence can be related to certain parts of the input context, we add a pointer generator~\cite{See2017GetTT} to the multimodal Transformer such that the rewrite is either generated from scratch or directly copied from the input. Specifically, the cross attention in the last decoder layer can be naturally taken as the context vector
\begin{equation}
    c_i = MultiHead(d_i,h^U_L,h^U_L)
\end{equation}
where $h^U_L$ is the textual part of the encoding result, which is the embedding of tokens before the ``[SEP]'' token. 
\begin{equation}
    \alpha_{t,i} = softmax(\frac{{(W_ss_t)}^TW_hh_i}{\sqrt{d}})
\end{equation}
\begin{equation}
    P_{copy}(y_t) = \sum_{i:w_i\in H}\alpha_{t,i}
\end{equation}
where $\alpha$ is the copy distribution over the input $H$, $P_{copy}$ determines where to copy at time step $t$.

By incorporating the pointer generator, the final probability of generating the target word $y_t$ at time step $t$ is represented as
\begin{equation}
    P(y_t) = \lambda P_{vocab}(y_t)+(1-\lambda)P_{copy}(y_t)
\end{equation}
\begin{equation}
    \lambda = sigmoid(w_d^Tc_t+w_l^Ts_t+w_a^Tx_{t})
\end{equation}
where $w_d, w_l, w_a$ are the weights to learn.
\subsubsection{Training}
We adopt the standard generation loss when fine-tuning the pre-trained model parameterized $\theta$ on our McQR task. At each time step $t$, the decoder output token is determined based on the generated rewrites of previous time steps denoted as $y_{<t}$, the input image and text encoding vector $e^v$ and $e^t$. We minimize the negative log-likelihood of generating the target rewrite $y$ given the image text input and previous generation results as 
\begin{equation}
    min -logP_\theta(y|e^t,e^v)=-\sum_{j=1}^{|y|}logP_\theta(y_t|e^t,e^v,y_{<t})
\end{equation}

\section{Experiments}
\subsection{Compared Methods}
We compare our methods with several baselines. 
\begin{itemize}
    \item Original~\cite{Elgohary2019CanYU} is the method where the rewrite is set to be the same as the input query.
    \item AllenNLP Coref~\cite{Gardner2018AllenNLPAD} is a deep learning based NLP tool. We utilize its  coreference resolution model to generate the rewrite.
    \item (L/T)-Gen~\cite{Su2019ImprovingMD} denotes the LSTM/Transformer based encoder-decoder generation model. For Transformer-based models, we report the performance of two variants: Early Fusion  which utilizes the same encoder to encode image and text features, and Late Fusion which first embeds image and text into vector spaces separately and then performs fusion into a joint embedding. 
    \item (L/T)-Ptr~\cite{See2017GetTT} adds a pointer generator to the (L/T)-Gen model.
    \item VL-(Bart/T5)~\cite{cho2021vlt5} is the mulitimodal implementation of large pre-trained language model Bart/T5.
    \item VL-(Bart/T5)-Ptr is the model depicted in Section \ref{model} that adds a pointer generator to the pre-trained model.
\end{itemize}

\subsection{Experimental Settings}
Our code is implemented based on Pytorch and Huggingface Transformers~\cite{Wolf2019HuggingFacesTS}. We use the base version of the pre-trained VL-Bart/T5 in all our experiments. Our dataset is split into the training, testing, and validation dataset following the portion of 60\%, 20\%, 20\%, resulting in 48566 queries for the training set, 16189 queries for the testing and validation set. By default, we set the batch size as 32 and the learning rate to be 5e-5, the model is fine-tuned for 20 epochs with the random seed of 42. In the testing stage, all models decode words by beam search with beam size set to 5.

We employ several evaluation metrics to evaluate the quality of the generated rewrite. We first report the BLEU~\cite{Papineni2002BleuAM}, ROUGE~\cite{Lin2004ROUGEAP}, and METEOR~\cite{Denkowski2014MeteorUL} rate. In addition, we report the Exact Match (EM) of both positive samples that involves some changes in the rewrite and negative samples that the rewrite is the same as the query. On two subtasks, we report the precision, recall, and F1 score.

\subsection{Experiment Results}
We first report the overall performance on the query rewrite task. Then we perform experiments on two subtasks including coreference resolution and ellipsis completion, respectively.

Table \ref{tab:mainexp} demonstrates the overall performance of different models on the McQR task. We have the following observations: first of all, compared with LSTM based models (e.g. L-Gen), Transformer based models have a stronger generation ability (e.g. T-(E) Gen), where the BLEU-2 score increases from 77.14 to 78.03. In addition, with the help of the pointer generator, certain parts from the original query and the history context are able to be preserved in the rewrite, where the negative EM score increases from 86.89 to 87.22 in the VLBart model. Furthermore, by leveraging the pre-trained model, a significant performance improvement can be seen where the BLEU-2 increases from 80.12 in T(L)-Ptr to 90.47 in VLT5-Ptr. 

\label{overall}
\begin{table}[]
    \centering
    \small
    \begin{tabular}{cccc}
    \hline
         & Precision & Recall & F1  \\
    \hline
        AllenNLP Coref & 15.76 & 14.48 &15.10 \\
    \hline
        L-Gen & 28.63 & 28.18 & 28.40 \\
        T-Gen & 42.12 & 41.45 & 41.78 \\
        L-Ptr & 38.07 & 37.47 & 37.77\\
        T-Ptr & 43.88 & 43.19 & 43.53\\ 
    \hline
        MBERT & 72.42 & 65.03 & 68.53\\
        VLT5 & 77.00 & 70.84 & 73.79\\
        VLBart & 76.59 & 70.37 & 73.35 \\
        VLBart-Ptr & 77.28 & 71.00 & 74.00\\
        VLT5-Ptr & \textbf{77.33} & \textbf{71.05} & \textbf{74.05}\\
        
    \hline
    \end{tabular}
    \caption{Precision, recall, F1 score of visual coreference resolution. For Transformer based models, we record the late fusion version. All three metrics are evaluated on utterance level.}
    \label{tab:coref}
\end{table}
\label{cr}
\begin{table}[]
    \centering
    \small
    \begin{tabular}{cccc}
    \hline
         & Precision & Recall & F1  \\
    \hline
        L-Gen & 17.95 & 17.29 &17.61\\
        T-Gen & 29.02 & 30.12 & 29.56\\
        L-Ptr & 24.49 & 23.59 & 24.03\\
        T-Ptr & 32.60 & 31.41 & 32.00 \\ 
    \hline
        VLBart & 51.93 & 50.04 & 50.97\\
        VLT5 & 54.33 & 52.35 & 53.32\\
        VLBart-Ptr & 52.45 & 50.54 & 51.47\\
        VLT5-Ptr & \textbf{54.59} & \textbf{52.60} &\textbf{53.58}\\
    \hline
    \end{tabular}
    \caption{Precision, recall, F1 score of visual ellipsis completion. For Transformer based models, we also report the late fusion version.}
    \label{tab:ellipsis}
\end{table}
 Table \ref{tab:coref} shows the results of different models on the visual coreference resolution (VCR) subtask. We further utilize one state-of-the-art VCR method named MBERT to compare~\cite{Yu2022VDPCRIV}.  According to the table, traditional text-only coreference resolution methods such as AllenNLP Coref may not have a good performance in the VCR task. The reason is that, in the conversations, many pronouns refer to entities that never appear in the history context, but can be found clues in the image. Furthermore, it can be concluded that pointer network also has a positive influence on the results since many coreferred entities can be directly copied from previous turns. 

Besides VCR, we also perform tests on the visual ellipsis completion task, as shown in Table \ref{tab:ellipsis}. The overall condition is similar to results in Table \ref{tab:coref}, while there encounters a performance degradation in this task. This may result from the fact that  recovering the omitted information may not be that apparent  (we will use some examples to illustrate in Section \ref{casestudy}). In conclusion, our model shows superior performance on both two tasks, demonstrating the effectiveness of our model in resolving the information omission of the conversations.

\label{ec}
\section{Extensive Analysis}
\subsection{Image Role Analysis}
\begin{figure}
    \centering
    \includegraphics[width=\linewidth]{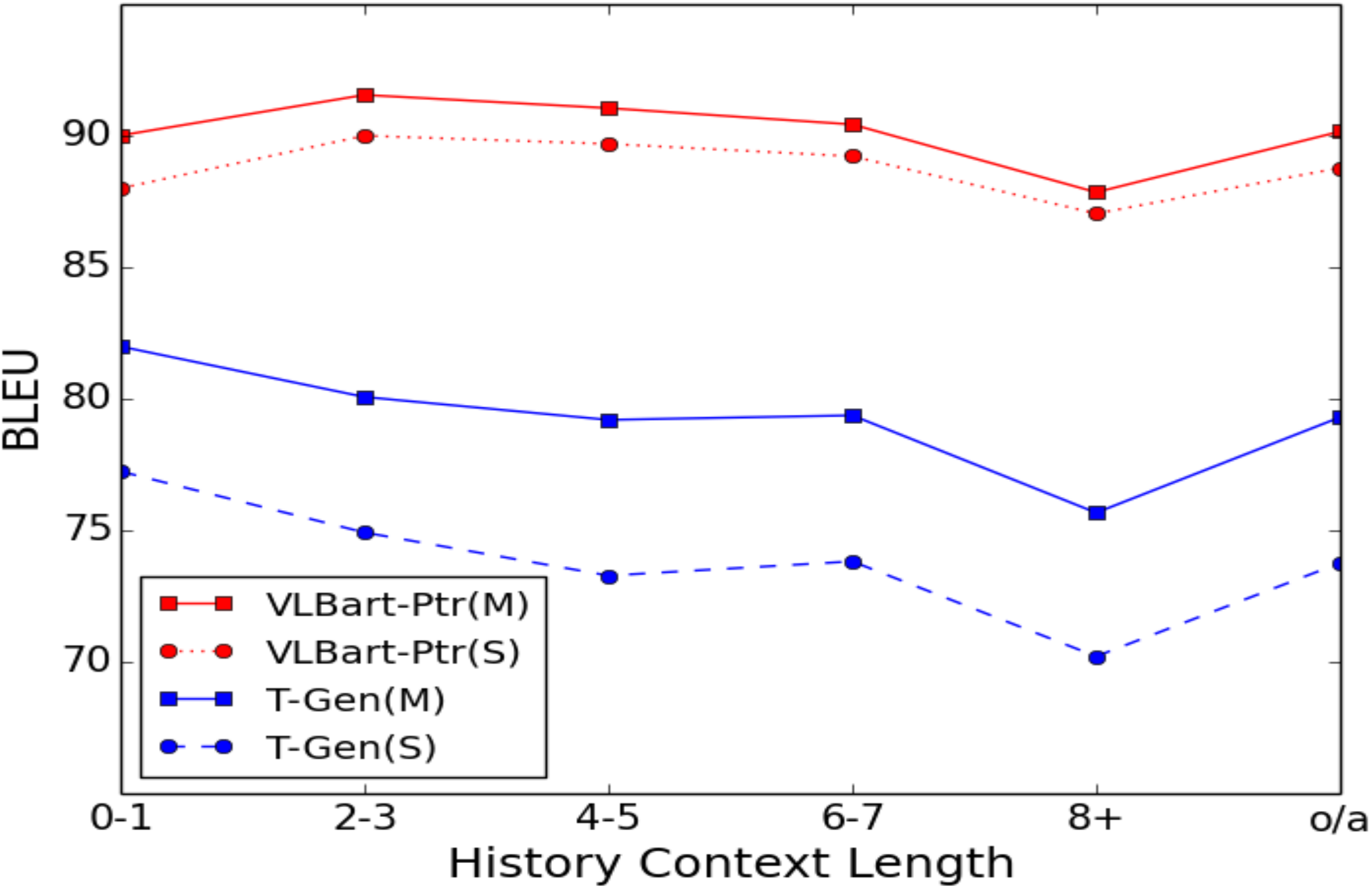}
    \caption{BLEU-2 performance under different length of turns, where (M) denotes the multimodal version, (S) denotes the text-only version, o/a denotes the overall performance over all turns.}
    \label{fig:turn}
\end{figure}
\begin{figure*}
    \centering
    \includegraphics[width=\linewidth]{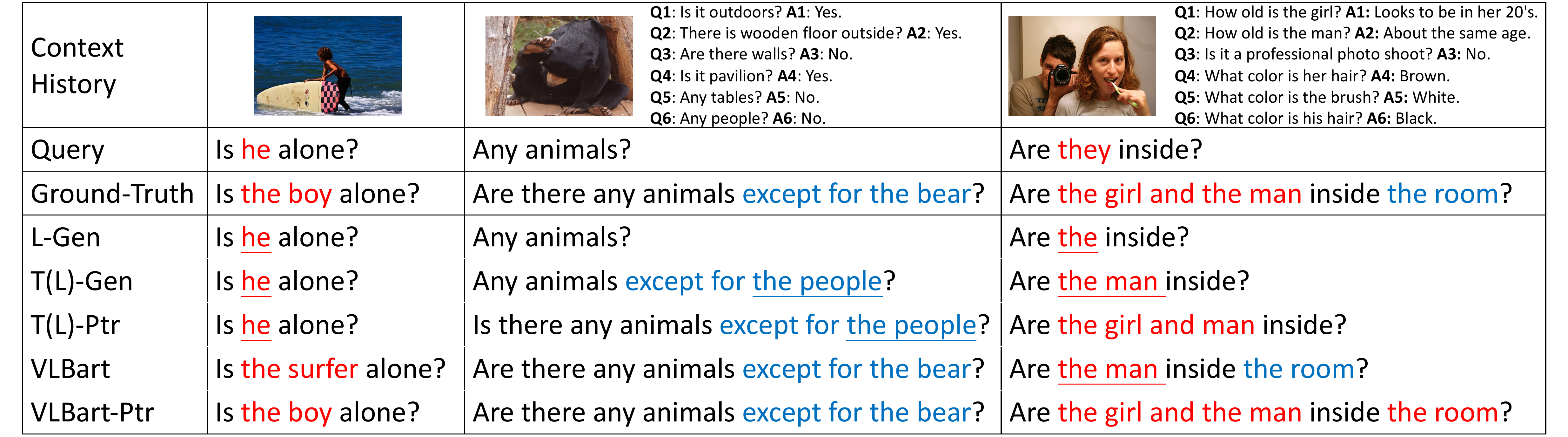}
    \caption{Example rewrites generated by different methods, where the red part denotes the coreference part and the blue part denotes the ellipsis part. The underline denotes the errors in the rewrites.}
    \label{fig:casestudy}
\end{figure*}

\begin{figure}
    \centering
    \includegraphics[width=\linewidth]{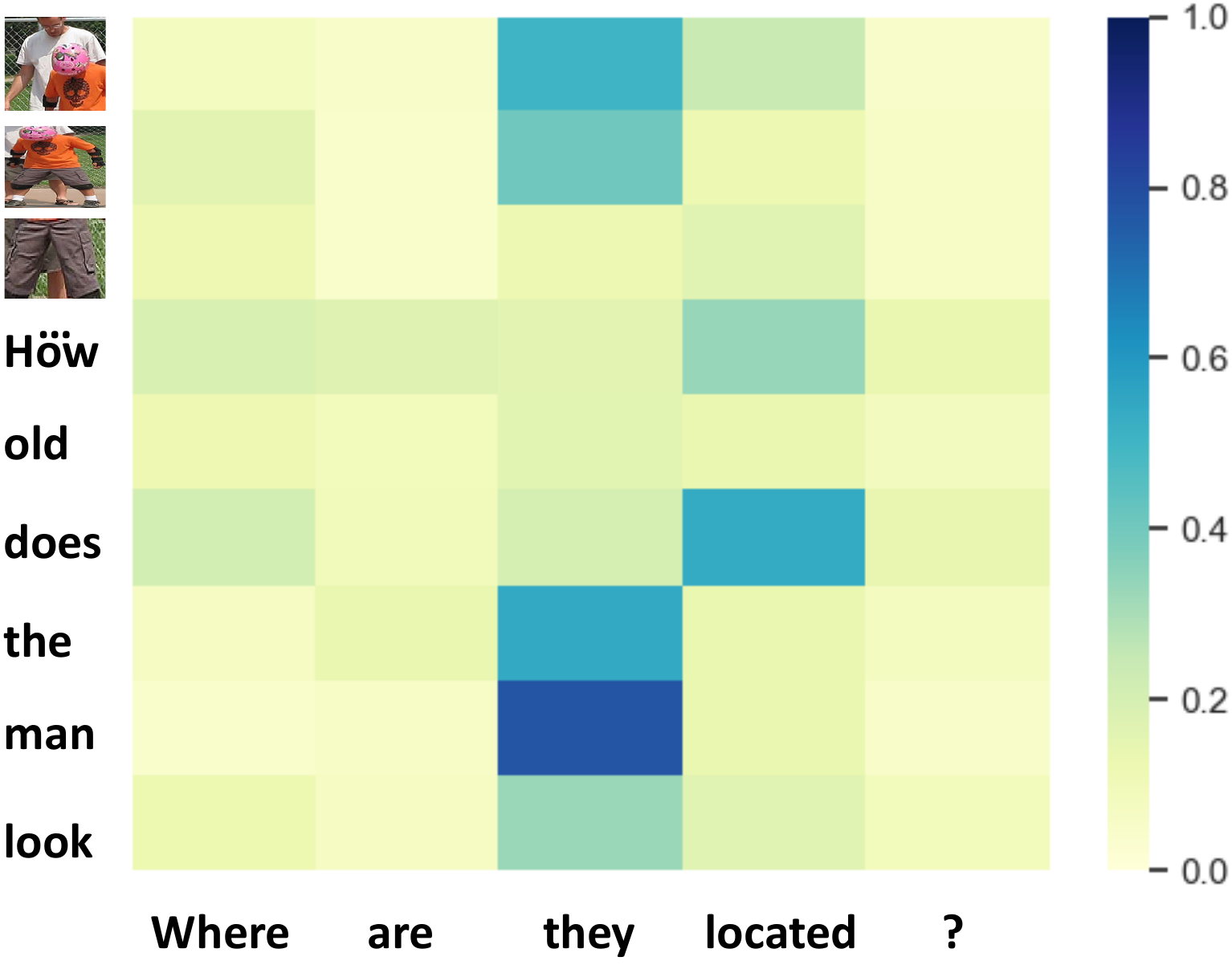}
    \caption{The heat map visualization of the self-attention weight in the Transformer block of our model. The full visual conversation example is shown in Appendix \ref{de}.}
    \label{fig:heat}
\end{figure}
In order to further investigate the role of image information in the query rewrite task,  we remove the image features and compare the performance with the original model under different length of context respectively. Specifically, we divide the overall 16189 testing data into five categories by the history context length, ending up with
3136/3380/3180/3185/3308 data records with the context turn length from 0-1/2-3/4-5/6-7/8+,  as shown in Figure \ref{fig:turn}.
According to the figure, the performance gets improved with the help of image information. Compared with Transformer generator, our model is less sensitive to the turn length, where the model still has a decent performance when the conversation goes deep. We can also observe that, in our model, when there is no or short textual context that the information in conversation history is limited, the gap between monomodal and multimodal performance reaches to the largest, showing that the model is more dependant to the image information in this case. While when the conversation is deep, the gap decreases, showing that the model learns to copy information from the rich history context when rewriting the query.

\subsection{Multimodal Attention Visualization}
We utilize the self-attention weight heat map in the Transformer block of our model to visualize how the model learns the cross-modal relationship of the conversation. In the example shown in Figure \ref{fig:heat} (the full conversation information can be found in Appendix \ref{de}), the first three image segments correspond to ROI ``man'', ``boy'', ``shorts'', respectively. According to the map, the pronoun ``they'' is correctly related to the ``man'' and ``boy'' ROI segments and also has a high attention weight to the text ``the man'' in the history context\footnote{The pronoun ``they'' in the query refers to ``the man and the boy'' in the original conversation.}. The learned relationship between visual and text representations can serve as the reason that our model shows superior performance concerning the multimodal feature incorporation in the McQR task.

\subsection{Case Study}
\label{casestudy}

We provide several example rewrites generated by different methods to vividly demonstrate the quality of rewrites, according to Figure \ref{fig:casestudy}.
In the first case where the query is the first utterance where no textual context is provided, non pre-trained models (e.g. T(L)-Ptr) have difficulty resolving the pronoun ``he'' in the query. The results imply that with the help of prior knowledge, multimodal pre-trained models have a superior ability in incorporating image and text features, mitigating the gap between representations of multimodal heterogeneous space. The second case shows that traditional Transformer pointer network tends to copy incorrect context segments whose rewrite results may not make sense to human readers. For example, the abbreviated entity after ``except for'' should apparently be ``the bear'' instead of ``the people'', while our model has a better ability of understanding the whole context and outputing more accurate results. Furthermore, when it comes to complicated cases where coreference and ellipsis are both observed, our model is both capable of copying the entities from history to solve the coreference (e.g. the red part in the third case) and generate parts that require 
reasoning (e.g. the blue part in the third case) from scratch. 
\begin{figure*}
    \centering
    \includegraphics[width=\linewidth]{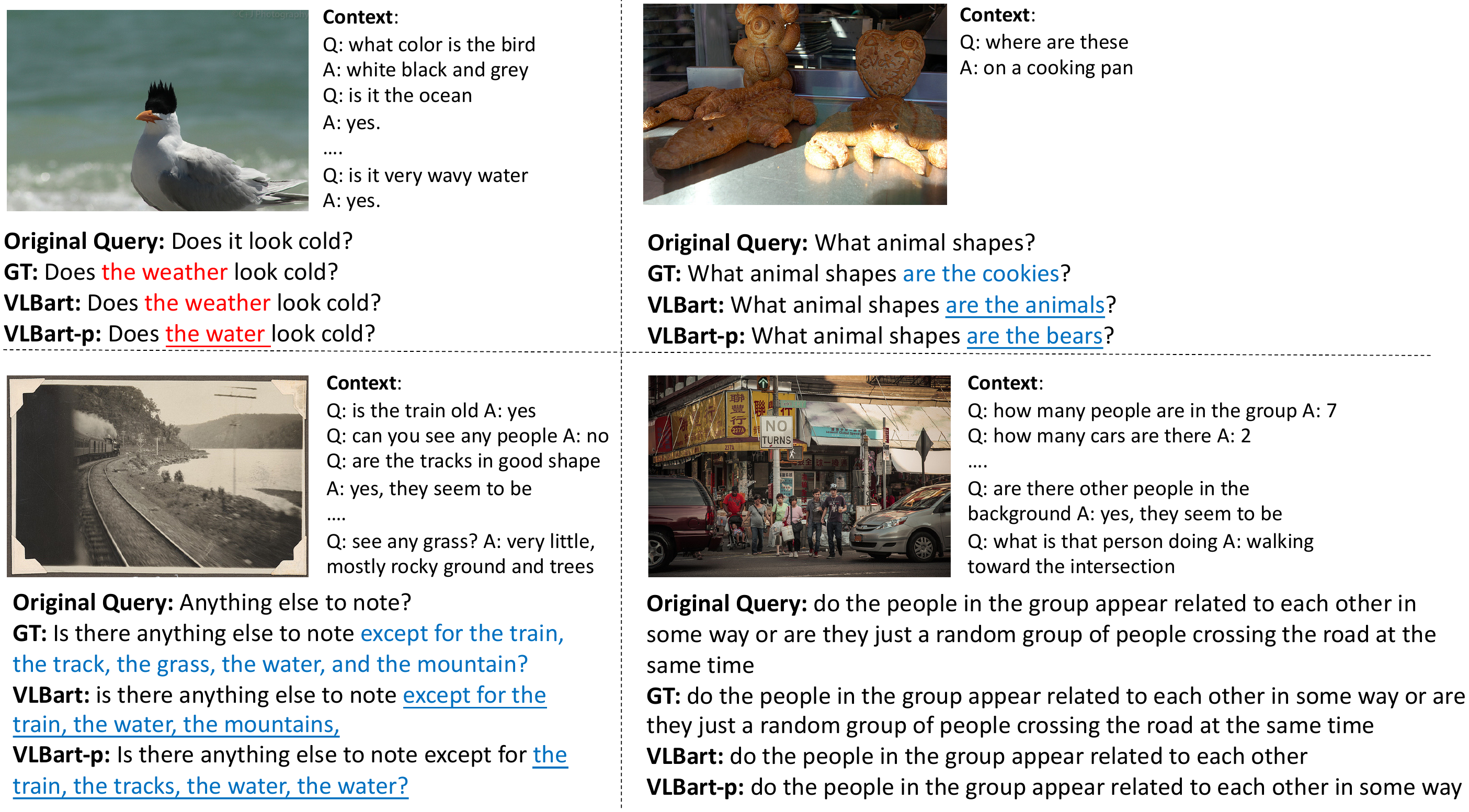}
    \caption{Common error cases of our method.}
    \label{fig:error}
\end{figure*}

\section{Error Case Analysis}
As shown in Figure \ref{fig:error}, we demonstrate the most common error cases of our model, which can be classified into several cases including bad sentence structure, wrong coreference result, inaccurate omission completion, etc. According to the figure, we can see that pre-trained language models can generate sentences that have a proper format, but still have some problems in understanding the deep semantic correlation in the visual conversations. For example, in the top two cases, although the coreference seems simple for human readers, it poses challenges for machines other than directly making a copy from context. Specifically, for complicated cases such as when coreference and ellipsis appear at the same time, when the conversation is long, and when the omitted information requires reasoning between history entities (e.g. the bottom cases), there still remains much space to explore. 

\section{Conclusion and Future Work}
We propose the task of multimodal conversational query rewrite (McQR), which aims to perform query rewrite under a multi-turn visual conversation. 
To facilitate the research, we benchmark a large-scale dataset with manual annotation which covers 15k visual conversations with more than 80k rewrites. We also provide image box annotation of entities appearing in the rewrites. Accordingly, we propose a model based on an existing multimodal pre-trained model and further enhance it with a pointer generator. Extensive experiments are performed to show the effectiveness of our model.

\section{Limitations}
Although our model achieves over 90\% BLEU-2 rate on our dataset, there is still room for improvements in the future work, since the exact match for positive rewrite queries can be further lifted. In addition, in the future work, we will also continue to explore how the dataset and model can further benefit downstream research under this scenario. Furthermore, fine-grained image features such as the image box annotation can be leveraged for improving the performance.

\section{Ethics Statement}
All data records of our dataset are originally from the Visual Dialog~\cite{Das2017VisualD} dataset, where all the images are collected from the COCO~\cite{Lin2014MicrosoftCC} dataset. When annotating the data, we make sure that the annotators do not have any other rights except for the conversation information. Upon data publication, we will strictly follow the user terms of the Visual Dialog dataset.



\bibliography{anthology,custom}
\appendix


\section{Rewrite Annotation Interface}
\label{interface}
\begin{figure}[H]
    \centering
    \includegraphics[width=\linewidth]{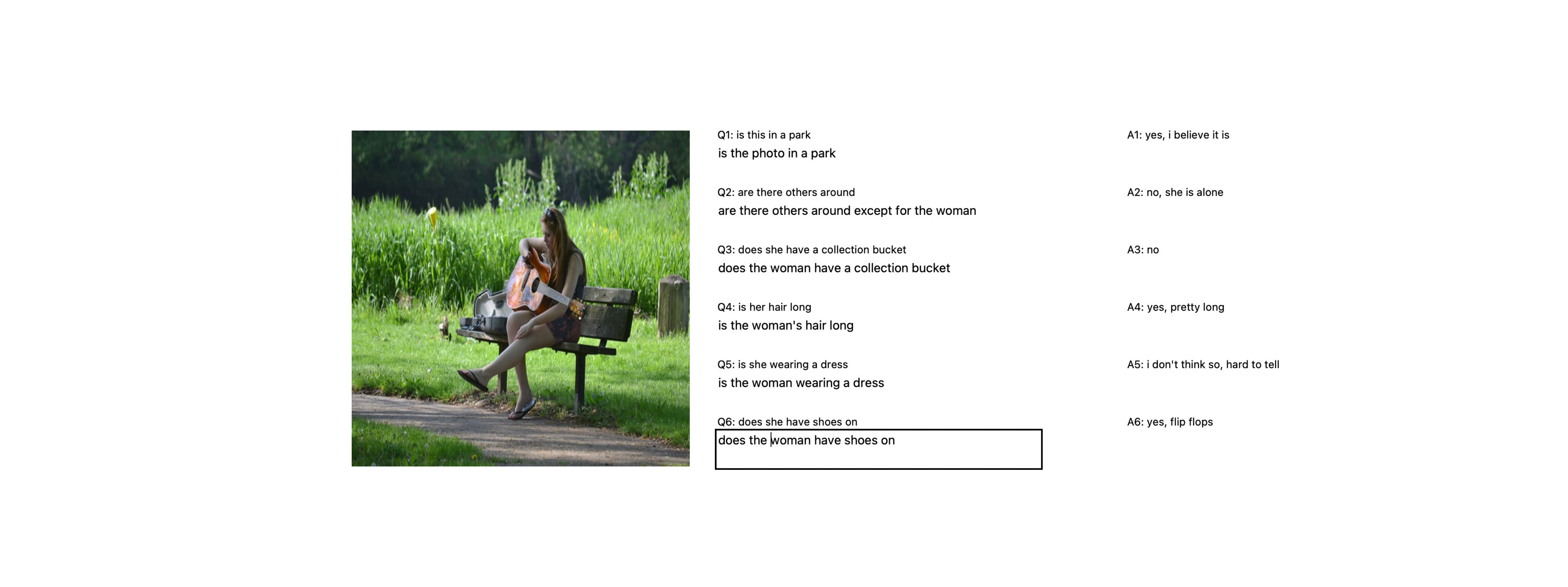}
    \caption{Rewrite annotation interface of our dataset.}
    \label{fig:interface1}
\end{figure}

Figure \ref{fig:interface1} shows the annotation interface when preparing the rewrite. We assign one text box for each query in the visual conversation so that the annotators can rewrite the query according to the image and history context. 

\section{Image Box Annotation}
\label{iba}
\begin{figure}[H]
    \centering
    \includegraphics[width=\linewidth]{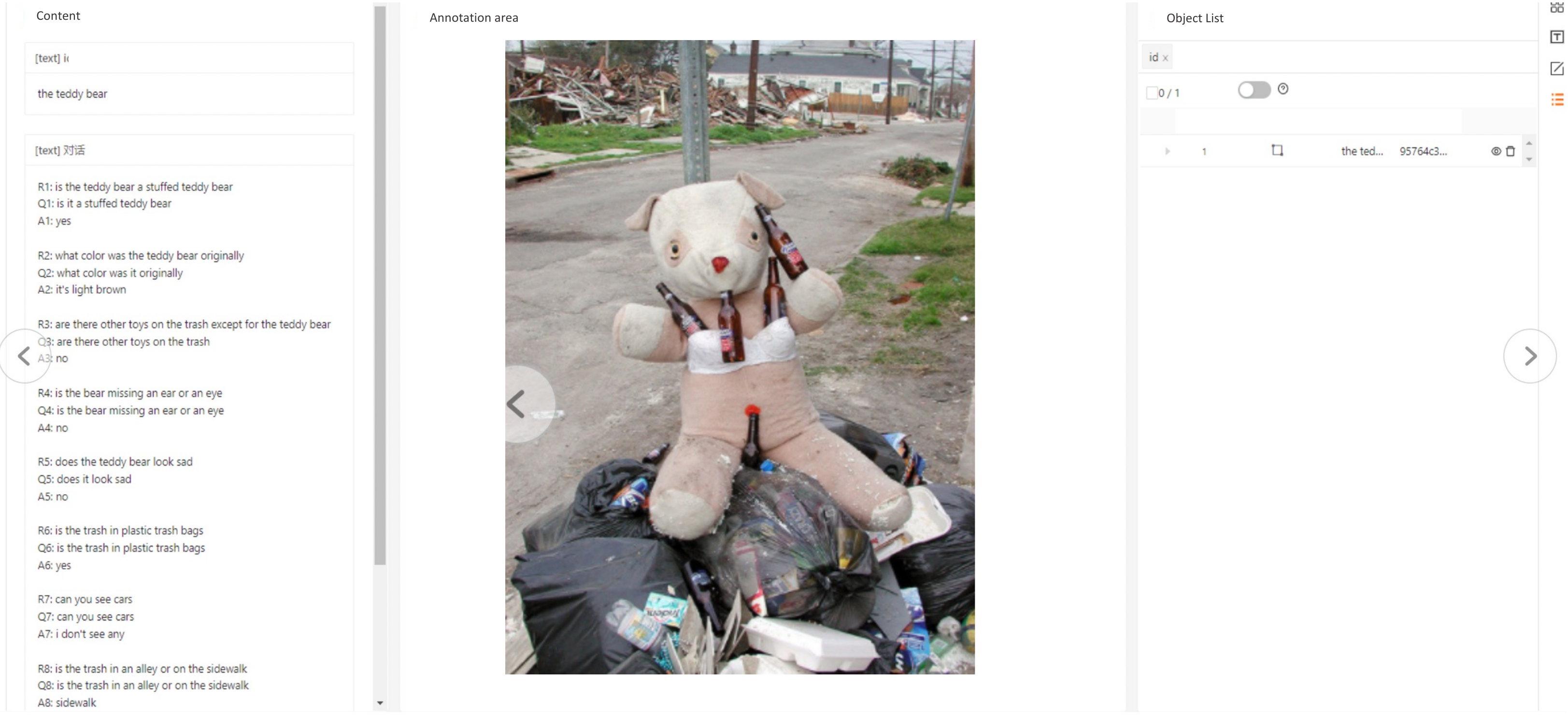}
    \caption{Image box annotation interface of our dataset.}
    \label{fig:interface2}
\end{figure}

Figure \ref{fig:interface2} shows the image box annotation interface, where the annotators extract entities from the rewrites and mark them on the image in the format of box.
\section{Quality Control Questionnaire}
\label{qc}
\begin{figure}[H]
    \centering
    \includegraphics[width=\linewidth]{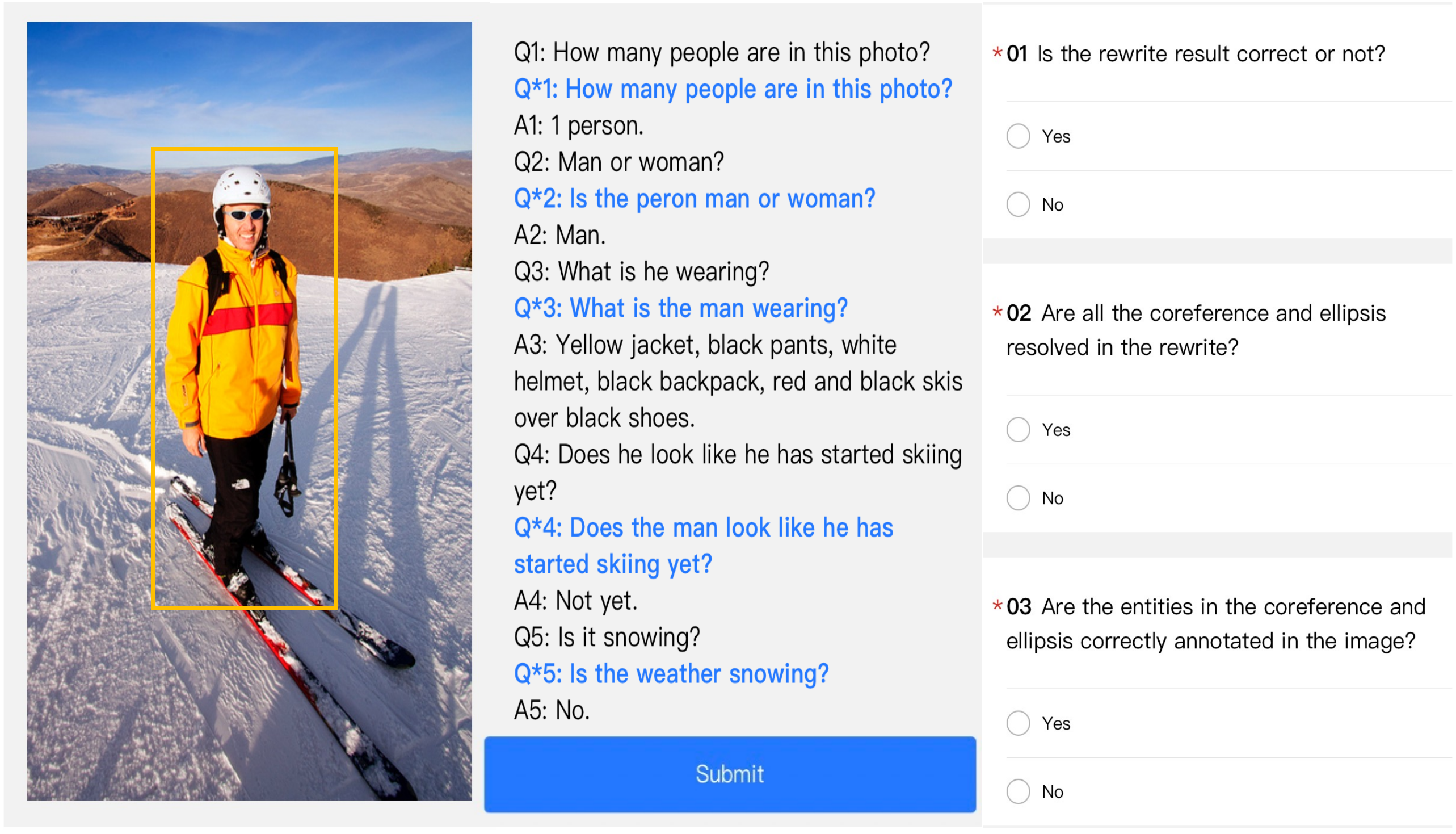}
    \caption{The questionnaire interface in the quality control phase.}
    \label{fig:questionnaire}
\end{figure}

We design the quality control questionnaire as shown in Figure \ref{fig:questionnaire}. All quality checkers have to click and answer the three questions.

\section{Dataset Examples}
\label{de}
\begin{figure}[H]
    \centering
    \includegraphics[width=\linewidth]{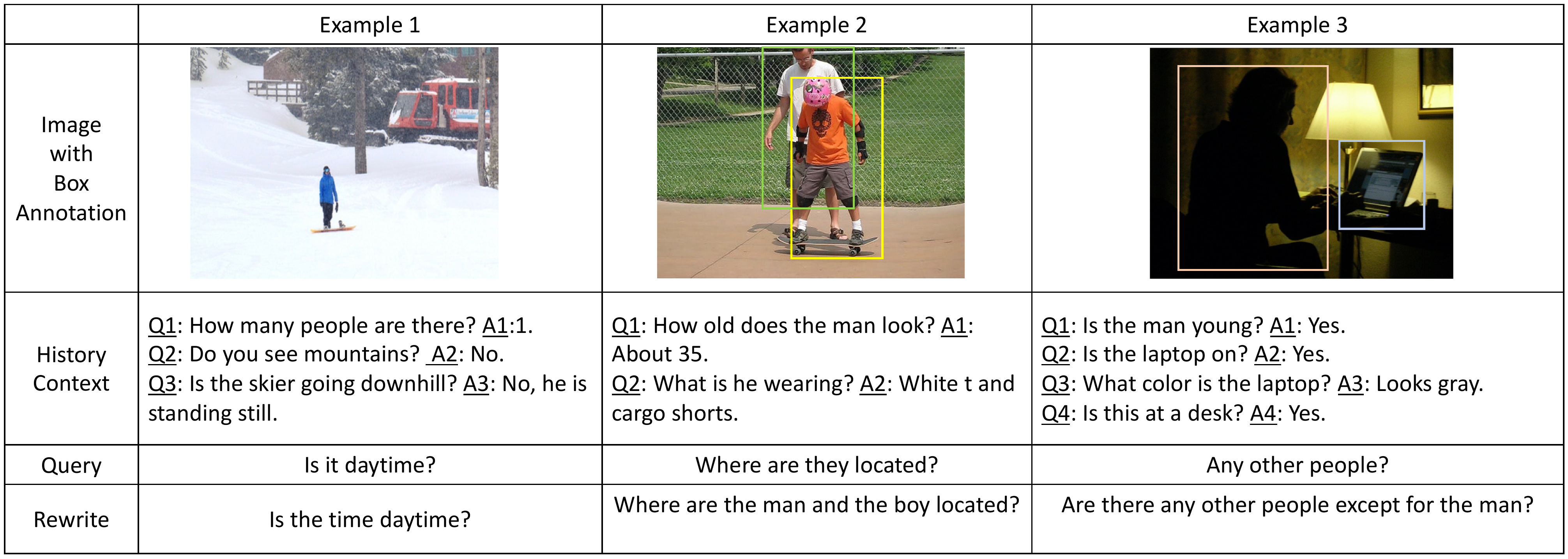}
    \caption{Example cases of our dataset.}
    \label{fig:de}
\end{figure}

We list some examples of our dataset to vividly demonstrate the characteristic of the task.  All the examples are listed in Figure \ref{fig:de}.
\end{document}